\documentclass[conference]{IEEEtran}
\usepackage{cite}
\usepackage{amsmath,amssymb,amsfonts}
\usepackage{algorithmic}
\usepackage{graphicx}
\usepackage{textcomp}
\usepackage{hyperref}
\usepackage{verbatim}

\hypersetup{final,hidelinks}
\ifCLASSINFOpdf
\else
\fi

\IEEEoverridecommandlockouts
\begin{document}
%

\title{Language Model-Based Paired Variational Autoencoders for Robotic Language Learning}


\author{Ozan \"Ozdemir, Matthias Kerzel, Cornelius Weber, Jae Hee Lee, Stefan Wermter
\thanks{O. \"Ozdemir, M. Kerzel, C. Weber, J. H. Lee and S. Wermter are with Knowledge Technology Group, Department of Informatics, University of Hamburg, Hamburg, Germany (emails: ozan.oezdemir@*, matthias.kerzel@*, cornelius.weber@*, jae.hee.lee@*, stefan.wermter@*; * = uni-hamburg.de)}}

\maketitle
\begin{abstract}
Human infants learn language while interacting with their environment in which their caregivers may describe the objects and actions they perform. Similar to human infants, artificial agents can learn language while interacting with their environment. In this work, first, we present a neural model that bidirectionally binds robot actions and their language descriptions in a simple object manipulation scenario. Building on our previous Paired Variational Autoencoders (PVAE) model, we demonstrate the superiority of the variational autoencoder over standard autoencoders by experimenting with cubes of different colours, and by enabling the production of alternative vocabularies. Additional experiments show that the model's channel-separated visual feature extraction module can cope with objects of different shapes. Next, we introduce PVAE-BERT, which equips the model with a pretrained large-scale language model, i.e., Bidirectional Encoder Representations from Transformers (BERT), enabling the model to go beyond comprehending only the predefined descriptions that the network has been trained on; the recognition of action descriptions generalises to unconstrained natural language as the model becomes capable of understanding unlimited variations of the same descriptions. Our experiments suggest that using a pretrained language model as the language encoder allows our approach to scale up for real-world scenarios with instructions from human users.
\end{abstract}


%
\IEEEpeerreviewmaketitle

\section{Introduction}
Humans use language as a means to understand and to be understood by their interlocutors. Although we can communicate effortlessly in our native language, language is a sophisticated form of interaction which requires comprehension and production skills. Understanding language depends also on the context, because words can have multiple meanings and a situation can be explained in many ways. As it is not always possible to describe a situation only in language or understand it only with the medium of language, we benefit from other modalities such as vision and proprioception. Similarly, artificial agents can utilise the concept of embodiment (i.e. acting in the environment) in addition to perception (i.e. using multimodal input like audio and vision) for better comprehension and production of language \cite{bisk2020experience}. Human infants learn language in their environment while their caregivers describe the properties of objects, which they interact with, and actions, which are performed on those objects. In a similar vein, artificial agents can be taught language; different modalities such as audio, touch, proprioception and vision can be employed towards learning language in the environment.

\begin{figure}[t]
    \centering
    \includegraphics[width=0.48\textwidth]{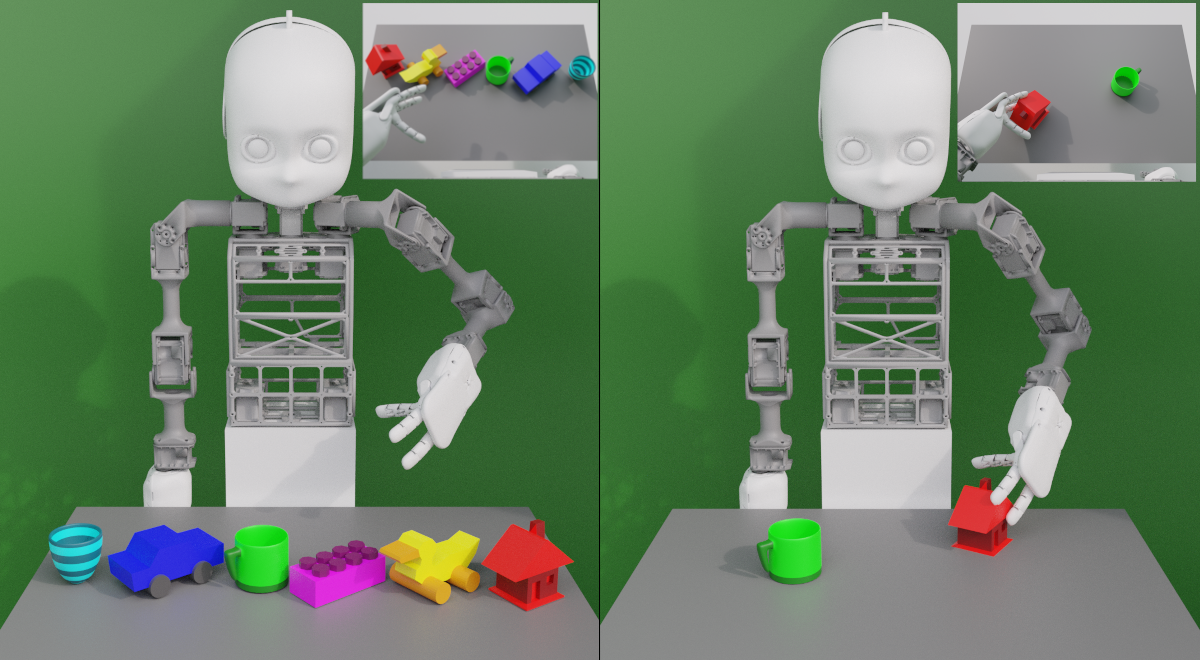}
    \caption[The NICO robot in the simulation]{Our tabletop object manipulation scenario in the simulation environment: the NICO robot is interacting with toy objects. In the left panel, NICO views all the toy objects; on the right, NICO pulls the red house. In both panels, NICO's field of view is given in the top right inset.}
    \label{fig:nico}
\end{figure}

The field of artificial intelligence has recently seen many studies attempting to learn language in an embodied fashion \cite{ng2017hey, heinrich2018interactive, chai2018language, lynch2021language, Akakzia_DECSTR_2021}. In this paper, we bidirectionally map language with robot actions by employing three distinct modalities, namely text, proprioception and vision. In our robotic scenario, two objects\footnote{Note that, in the left panel of Fig. \ref{fig:nico}, we show all the toy objects for visualisation purposes. In all our experiments, there are always only two objects on the table.} are placed on a table as the NICO (the Neuro-Inspired COmpanion) robot \cite{kerzel2017nico} physically interacts with them - see Figure \ref{fig:nico}. NICO moves objects along the table surface according to given textual descriptions and recognises the actions by translating them to corresponding descriptions. The possibility of bidirectional translation between language and control was realised with a paired recurrent autoencoder (PRAE) architecture by Yamada et al. \cite{yamada2018paired}, which aligns the two modalities that are each processed by an autoencoder. We extended this approach (PRAE) with the Paired Variational Autoencoders (PVAE) \cite{Oezdemir_2021_ICDL} model, which enriches the language used to describe the actions taken by the robot: instead of mapping a distinct description to each action \cite{yamada2018paired}, the PVAE maps multiple descriptions, which are equivalent in meaning, to each action. Hence, we have transcended the strict one-to-one mapping between control and language since our variational autoencoder-based model can associate each robot action with multiple description alternatives. The PVAE is composed of two variational autoencoders (VAEs), one for language, the other for action, and both of them consist of an LSTM (long short-term memory) \cite{hochreiter1997long} encoder and decoder which are suitable for sequential data. The dataset\footnote{\url{https://www.inf.uni-hamburg.de/en/inst/ab/wtm/research/corpora.html}}, which our model is trained with, consists of paired textual descriptions and corresponding joint angle values with egocentric images. The language VAE reconstructs descriptions, whereas the action VAE reconstructs joint angle values that are conditioned on the visual features extracted in advance by the channel-separated convolutional autoencoder (CAE) \cite{Oezdemir_2021_ICDL} from egocentric images. The two autoencoders are implicitly bound together with an extra loss term which aligns actions with their corresponding descriptions and separates unrelated actions and descriptions in the hidden vector space.

However, even with multiple descriptions mapped to a robot action as implemented in our previous work \cite{Oezdemir_2021_ICDL}, replacing each word by its alternative does not lift the grammar restrictions on the language input. In order to process unconstrained language input, we equip the PVAE architecture with the Bidirectional Encoder Representations from Transformers (BERT) language model \cite{devlin2019bert} that has been pretrained on large-scale text corpora to enable the recognition of unconstrained natural language commands by human users. To this end, we replace the LSTM language encoder with a pretrained BERT model so that the PVAE can recognise different commands that correspond to the same actions as the predefined descriptions given the same object combinations on the table. This new model variant, which we call PVAE-BERT, can handle not only the descriptions it is trained with, but also various descriptions equivalent in meaning with different word order and/or filler words (e.g., `please', `could', `the', etc.) as our analysis shows. We make use of transfer learning by using a pretrained language model, hence, benefitting from large unlabelled textual data.

Our contributions can be summarised as follows:
\begin{enumerate}
\item In our previous work \cite{Oezdemir_2021_ICDL}, we showed that variational autoencoders facilitate better one-to-many action-to-language translation and that channel separation in visual feature extraction, i.e., training RGB channels separately, results in more accurate recognition of object colours in our object manipulation scenario. In this follow-up work, we extend our dataset with different shapes and show that our PVAE with the channel separation approach is able to translate from action to language while manipulating different objects. 
\item Here, we introduce PVAE-BERT, which, by using pretrained BERT, indicates the potential of our approach to be scaled up for unconstrained instructions from human users.
\item Additional principle component (PCA) analysis shows that language as well as action representation vectors arrange according to the semantics of the language descriptions.
\end{enumerate}

The remainder of this paper is organised as follows: the next section describes the relevant work, Section 3 presents the architecture of the PVAE and PVAE-BERT models, various experiments and their results are given in Section 4, Section 5 discusses the results and their implications and the last section concludes the paper with final remarks\footnote{Our code is available at \url{https://github.com/oo222bs/PVAE-BERT}.}.

\section{Related Work}
The state-of-the-art approaches in embodied language learning mostly rely on tabletop environments \cite{hatori2018interactively, shridhar2018interactive, yamada2018paired, heinrich2020, shao2020concept2robot} or interactive play environments \cite{lynch2021language} where a robot interacts with various objects according to given instructions. We categorise these approaches into three groups: those that translate from language to action, those that translate from action to language and those that can translate in both directions, i.e., bidirectional approaches. Bidirectional approaches allow greater exploitation of available training data as training in both directions can be interpreted as multitask learning, which ultimately leads to more robust and powerful models independent of the translation direction. By using the maximum amount of shared weights for multiple tasks, such models would be more efficient than independent unidirectional networks in terms of data utilisation and the model size.

\subsection{Language-to-Action Translation}
Translating from language to action is the most common form in embodied language learning. Hatori et al. \cite{hatori2018interactively} introduce a neural network architecture for moving objects given the visual input and language instructions, as their work focuses on the interaction of a human operator with the computational neural system that picks and places miscellaneous items as per verbal commands. In their scenario, many items with different shape and size (e.g. toys, bottles etc.) are distributed across four bins with many of them being occluded - hence, the scene is very complex and cluttered. Given a pick-and-place instruction from the human operator, the robot first confirms and then executes it if the instruction is clear. Otherwise, the robot asks the human operator to clarify the desired object. The network receives a verbal command from the operator and an RGB image from the environment, and it has separate object recognition and language understanding modules, which are trained jointly to learn the names and attributes of the objects. 

Shridhar and Hsu \cite{shridhar2018interactive} propose a comprehensive system for a robotic arm to pick up objects based on visual and linguistic input. The system consists of multiple modules such as manipulation, perception and a neural network architecture, and is called INGRESS (Interactive Visual Grounding of Referring Expressions). INGRESS is composed of two network streams (self-referential and relational) which are trained on large datasets to generate a definitive expression for each object in the scene based on the input image. The generated expression is compared with the input expression to detect the desired object. INGRESS is therefore responsible for grounding language by learning object names and attributes via manipulation. The approach can resolve ambiguities when it comes to which object to lift by asking confirmation questions to the user. 

Shao et al. \cite{shao2020concept2robot} put forward a robot learning framework, Concept2Robot, for learning manipulation concepts from human video demonstrations in two stages. In the first stage, they use reinforcement learning and, in the second, they utilise imitation learning. The architecture consists of three main parts: semantic context network, policy network and action classification. The model receives as input a natural language description for each task alongside an RGB image of the initial scene. In return, it is expected to produce the parameters of a motion trajectory to accomplish the task in the given environment. 

Lynch and Sermanet \cite{lynch2021language} introduce the LangLfP (language learning from play) approach, in which they utilise multicontext imitation to train a single policy based on multiple modalities. Specifically, the policy is trained on both image and language goals and this enables the approach to follow natural language instructions during evaluation. During training, fewer than 1\% of the tasks are labelled with natural language instructions, because it suffices to train the policy for more than 99\% of the cases with goal images only. Therefore, only few of the tasks must be labelled with language instructions. Furthermore, they utilise a Transformer-based \cite{vaswani2017attention} multilingual language encoder, Multilingual Universal Sentence Encoder \cite{yang-etal-2020-multilingual}, to encode linguistic input so that the system can handle unseen language input like synonyms and instructions in 16 different languages.

The language-to-action translation methods are designed to act upon a given language input as in textual or verbal commands. They can recognise commands and execute the desired actions. However, they cannot describe the actions that they perform.

\subsection{Action-to-Language Translation}
Another class of approaches in embodied language learning translates action into language. Heinrich et al. \cite{heinrich2020} introduce an embodied crossmodal neurocognitive architecture, the adaptive multiple timescale recurrent neural network (adaptive MTRNN), which enables the robot to acquire language by listening to commands while interacting with objects in a playground environment. The approach has auditory, sensorimotor and visual perception capabilities. Since neurons at multiple timescales facilitate the emergence of hierarchical representations, the results indicate good generalisation and hierarchical concept decomposition within the network. 

Eisermann et al. \cite{ELWW21} study the problem of compositional generalisation, in which they conduct numerous experiments on a tabletop scenario where a robotic arm manipulates various objects. They utilise a simple LSTM-based network to describe the actions performed on the objects in hindsight - the model accepts visual and proprioceptive input and produces textual descriptions. Their results show that with the inclusion of proprioception as input and using more data in training, the network's performance on compositional generalisation improves significantly.

Similar to the language-to-action translation methods, the action-to-language translation methods work only in one direction: they describe the actions they perform in the environment. However, they are unable to execute a desired action given by the human user. Nevertheless, from the robotics perspective, it is desirable to have models that can also translate from action to language and not just execute verbal commands; such robots can explain their actions by verbalising an ongoing action, which also paves the way for more interpretable systems.

\subsection{Bidirectional Translation}
Very few embodied language learning approaches are capable of flexibly translating in both directions, hence, bidirectional. While unidirectional approaches are feasible for smaller datasets, we aim to research architectures that can serve as large-scale multimodal foundation models and solve multiple tasks in different modalities. By generating a discrete set of words, bidirectional models can also provide feedback to a user about the information contained within its continuous variables. By providing rich language descriptions, rather than only performing actions, such models can contribute to explainable AI (XAI) for non-experts. For a comprehensive overview of the field of XAI, readers can refer to the survey paper by Adadi and Berrada \cite{surveyxai}. 

In one of the early examples of bidirectional translation, Ogata et al. \cite{ogata2007parametricbias} present a model that is aimed at articulation and allocation of arm movements by using a parametric bias to bind motion and language. The method enables the robot to move its arms according to given sentences and to generate sentences according to given arm motions. The model shows generalisation towards motions and sentences that it has not been trained with. However, it fails to handle complex sentences. 

Antunes et al. \cite{antunes2019multitimescale} introduce the multiple timescale long short-term memory (MT-LSTM) model in which the slowest layer establishes a bidirectional connection between action and language. The MT-LSTM consists of two components, namely language and action streams, each of which is divided into three layers with varying timescales. The two components are bound by a slower meaning layer that allows translation from action to language and vice versa. The approach shows limited generalisation capabilities. 

Yamada et al. \cite{yamada2018paired} propose the paired recurrent autoencoder (PRAE) architecture, which consists of two autoencoders, namely action and description. The action autoencoder takes as input joint angle trajectories with visual features and is expected to reconstruct the original joint angle trajectories. The description autoencoder, on the other hand, reads and then reconstructs the action descriptions. The dataset that the model is trained on consists of pairs of simple robot actions and their textual descriptions, e.g., `pushing away the blue cube'. The model is trained end-to-end, with both autoencoders, reconstructing language and action, whilst there is no explicit neural connection between the two. The crossmodal pairing between action and description autoencoders is supplied with a loss term that aligns the hidden representations of paired actions and descriptions. The binding loss allows the PRAE to execute actions given instructions as well as translate actions to descriptions. As a bidirectional approach, the PRAE is biologically plausible to some extent, since humans can easily execute given commands and also describe these actions linguistically. To imitate human-like language recognition and production, bidirectionality is essential. However, due to its use of standard autoencoders, the PRAE can only bind a robot action with a particular description in a one-to-one way, although actions can be expressed in different ways. In order to map each robot action to multiple description alternatives, we have proposed the PVAE (paired variational autoencoders) approach \cite{Oezdemir_2021_ICDL} which utilises variational autoencoders (VAEs) to randomise the latent representation space and thereby allows one-to-many translation between action and language. A recent review by Marino \cite{marino2021predictive} highlights similarities between VAEs and predictive coding from neuroscience in terms of model formulations and inference approaches.

This work is an extension of the ICDL article ``Embodied Language Learning with Paired Variational Autoencoders'' \cite{Oezdemir_2021_ICDL}. Inspired by the TransferLangLfP paradigm by Lynch and Sermanet \cite{lynch2021language}, we propose to use the PVAE with a pretrained BERT language model \cite{devlin2019bert} in order to enable the model to comprehend unconstrained language instructions from human users. Furthermore, we conduct experiments using PVAE-BERT on our dataset for various use cases and analyse the internal representations for the first time.

\section{Proposed Methods: PVAE \& PVAE-BERT}

\begin{figure*}[ht]
    \centering
    \includegraphics[width=1\textwidth]{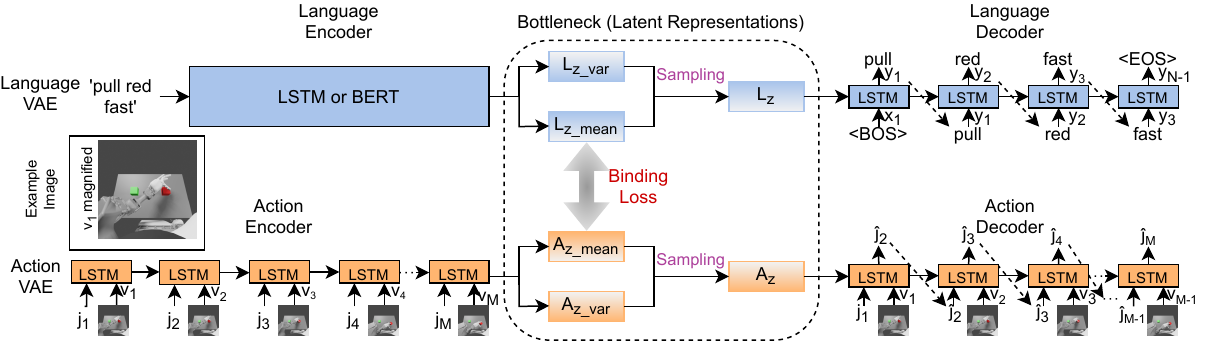}
    \caption[The architecture of the proposed PVAE and PVAE-BERT]{The architecture of the proposed PVAE and PVAE-BERT models: the language VAE (blue rectangles) processes descriptions, whilst the action VAE (orange rectangles) processes joint angles and images at each time step. The input to the language VAE is the given description $x$, whereas the action VAE takes as input joint angle values $j$ and visual features $v$. The two VAEs are implicitly bound via a binding loss in the latent representation space. The image from which the v\textsubscript{1} is extracted is magnified for visualisation purposes. $<$BOS$>$ and $<$EOS$>$ stand for \textsl{beginning of sentence} and \textsl{end of sentence} tags, respectively. The two models differ only by the language encoder employed: the PVAE uses LSTM, whereas PVAE-BERT uses a pretrained BERT model.}
    \label{fig:architecture}
\end{figure*}

As can be seen in Figure \ref{fig:architecture}, the PVAE model consists of two variational autoencoders: a language VAE and an action VAE. The former learns to generate descriptions matching original descriptions, whilst the latter learns to reconstruct joint angle values with conditioning on the visual input. The two autoencoders do not have any explicit neural connection between them, but instead they are implicitly aligned by the binding loss, which brings the two autoencoders closer to each other in the latent space over the course of learning by reducing the distance between the two latent variables. First, action and language encoder map the input to the latent code, i.e., the language encoder accepts one-hot encoded descriptions word by word as input and produces the encoded descriptions, whereas the action encoder accepts corresponding arm trajectories and visual features as input and produces the encoded actions. Next, the encoded representations are used to extract latent representations by randomly sampling from a Gaussian distribution separately for language and action modalities. Finally, from the latent representations, language and action decoders reconstruct the descriptions and joint angle values, respectively.

Our model is a bidirectional approach, i.e., after training translation is possible in both directions, action-to-language and language-to-action. The PVAE model transforms robot actions to descriptions in a one-to-many fashion by appropriately randomising the latent space. PVAE-BERT additionally handles variety in language input by using pretrained BERT as the language encoder module.
As part of the action encoder, the visual input features are extracted in advance using a channel-separated CAE (short for convolutional autoencoder), which improves the ability of the approach to distinguish the colours of cubes. The details of each model component are given in the following subsections.

\subsection{Language Variational Autoencoder}
The language VAE accepts as input one-hot encoded matrix of a description word by word in the case of the PVAE or the complete description altogether for PVAE-BERT, and for both the PVAE and PVAE-BERT, is responsible for reproducing the original description. It consists of an encoder, a decoder and latent layers (in the bottleneck) where latent representations are extracted via sampling. For the PVAE, the language encoder embeds a description of length $N$, $(x_{1}, x_{2}, ..., x_{N})$, into two fixed-dimensional vectors $z_{\text{mean}}$ and $z_{\text{sigma}}$ as follows:
\begin{align*}
h_{t}^{\text{enc}}, c_{t}^{\text{enc}} &= \text{LSTM}(x_{t}, h_{t-1}^{\text{enc}}, c_{t-1}^{\text{enc}}) \hspace{5mm} (1\leq t\leq N),\\
z_{\text{mean}}&= W^{\text{enc}}_{\text{mean}} \cdot h_{N} + b^{\text{enc}}_{\text{mean}}, \\
z_{\text{var}}&= W^{\text{enc}}_{\text{var}} \cdot h_{N} + b^{\text{enc}}_{\text{var}}, \\
z_{\text{lang}} &= z_{\text{mean}} + z_{\text{var}} \cdot \mathcal{N}(\mu, \sigma^{2}),
\end{align*}
where 
$h_{t}$ and $c_{t}$ are the hidden and cell state of the LSTM at time step $t$, respectively, and $\mathcal{N}$ is a Gaussian distribution. $h_{0}$ and $c_{0}$ are set as zero vectors, while $\mu$ and $\sigma$ are 0 and 0.1, respectively. $z_{\text{lang}}$ is the latent representation of a description. $\text{LSTM}$ here, and in the following, is a peephole LSTM \cite{sak2014long} following the implementation of Yamada et al. \cite{yamada2018paired}. The language input is represented in one-hot encoded matrices, whose rows represent the sequence of input words and columns represent every word that is in the vocabulary. In each row, only one cell is 1 and the rest are 0, which determines the word that is given to the model at that time step.

For PVAE-BERT, we replace the LSTM language encoder with the pretrained BERT-base model and, following the implementation by Devlin et al. \cite{devlin2019bert}, tokenise the descriptions accordingly with the subword-based tokeniser WordPiece \cite{wu2016google}.

The language decoder generates a sequence by recursively expanding $z_{\text{lang}}$:
\begin{align*}
    h_{0}^{\text{dec}}, c_{0}^{\text{dec}} &= W^{\text{dec}} \cdot  z_{\text{lang}} + b^{\text{dec}}, \\
    h_{t}^{\text{dec}}, c_{t}^{\text{dec}} &= \text{LSTM}(y_{t-1}, h_{t-1}^{\text{dec}}, c_{t-1}^{\text{dec}}) \hspace{3mm} (1\leq t\leq N-1), \\
    y_{t} &= \text{soft}( W^{\text{out}} \cdot h_{t}^{\text{dec}} + b^{\text{out}}) \hspace{5mm} (1\leq t\leq N-1),
\end{align*}
where 
$\text{soft}$ denotes the softmax activation function. $y_{0}$ is the first symbol indicating the beginning of the sentence, hence the $<$BOS$>$ tag. 
\subsection{Action Variational Autoencoder}
The action VAE accepts a sequence of joint angle values and visual features as input and it is responsible for reconstructing the joint angle values. Similar to the language VAE, it is composed of an encoder, a decoder and latent layers (in the bottleneck) where latent representations are extracted via sampling. The action encoder encodes a sequence of length $M$, $((j_{1},v_{1}), (j_{2}, v_{2}), ..., (j_{M}, v_{M}))$, which includes concatenation of joint angles $j$ and visual features $v$. Note that the visual features are extracted by the channel-separated convolutional autoencoder beforehand. The equations that define the action encoder are as follows\footnote{For the sake of clarity, we use mostly the same symbols in the equations as in the equations of the language VAE.}:
\begin{align*}
h_{t}^{\text{enc}}, c_{t}^{\text{enc}} &= \text{LSTM}(v_{t}, j_{t}, h_{t-1}^{\text{enc}}, c_{t-1}^{\text{enc}}) \hspace{5mm} (1\leq t\leq M), \\
z_{\text{mean}} &= W^{\text{enc}}_{\text{mean}} \cdot h_{M} + b^{\text{enc}}_{\text{mean}}, \\
z_{\text{var}} &= W^{\text{enc}}_{\text{var}} \cdot h_{M} + b^{\text{enc}}_{\text{var}}, \\
z_{\text{act}} &= z_{\text{mean}} + z_{\text{var}} \cdot \mathcal{N}(\mu, \sigma^{2}),
\end{align*}
where 
$h_{t}$ and $c_{t}$ are the hidden and cell state of the LSTM at time step $t$, respectively, and $\mathcal{N}$ is a Gaussian distribution. $h_{0}$, $c_{0}$ are set as zero vectors, while $\mu$ and $\sigma$ are set as 0 and 0.1, respectively. $z_{\text{act}}$ is the latent representation of a robot action.

The action decoder reconstructs the joint angles:
\begin{align*}
    h_{0}^{\text{dec}}, c_{0}^{\text{dec}} &= W^{\text{dec}} \cdot  z_{\text{act}} + b^{\text{dec}},\\
    h_{t}^{\text{dec}}, c_{t}^{\text{dec}} &= \text{LSTM}(v_{t}, \hat{\jmath}_{t}, h_{t-1}^{\text{dec}}, c_{t-1}^{\text{dec}})\hspace{5mm}(1\leq t\leq M-1), \\
    \hat{\jmath}_{t+1} &= \text{tanh}( W^{\text{out}} \cdot h_{t}^{\text{dec}} + b^{\text{out}}) \hspace{5mm} (1\leq t\leq M-1),
\end{align*}
where 
$\text{tanh}$ denotes the hyperbolic tangent activation function and $\hat{\jmath}_{1}$ is equal to ${j}_{1}$, i.e. joint angle values at the initial time step.

\subsection{Visual Feature Extraction}
We utilise a convolutional autoencoder architecture, following Yamada et al. \cite{yamada2018paired}, to extract the visual features of the images. Different from the approach used in \cite{yamada2018paired}, we change the number of input channels the model accepts from three to one and train an instance of CAE for each colour channel (red, green and blue) to recognise different colours more accurately: channel separation. Therefore, we call our visual feature extractor the channel-separated CAE. The idea behind the channel-separated CAE is similar to depthwise separable convolutions \cite{chollet2017xception}, where completely separating cross-channel convolutions from spatial convolutions leads to better results in image classification as the network parameters are used more efficiently. The channel-separated CAE accepts a colour channel of $120 \times 160$ RGB images captured by the cameras in the eyes of NICO - referred also as the egocentric view of the robot - at a time. As can be seen in detail in Table \ref{tab:detailed}, it consists of a convolutional encoder, a fully-connected bottleneck (incorporates hidden representations) and a deconvolutional decoder. After training for each colour channel, we extract the visual features of each image for every channel from the middle layer in the bottleneck (FC 3). The visual features extracted from each channel are then concatenated to make up the ultimate visual features $v$.

\begin{table}[ht]
\caption{Detailed Architecture of Channel-Separated CAE}\label{tab:detailed}
 \renewcommand{\tabcolsep}{2.5pt}
\begin{tabular}[t]{|c|c|c|c|c|c|c|}
 \hline
 Block & Layer & Out Chan. & Kernel Size & Stride & Padding & Activation\\
 \hline
 Encoder & Conv 1 & 8 & 4x4 & 2 & 1 & ReLU\\
  & Conv 2 & 16 & 4x4 & 2 & 1 & ReLU\\
& Conv 3 & 32 & 4x4 & 2 & 1 & ReLU\\
& Conv 4 & 64 & 8x8 & 5 & 2 & ReLU\\
 \hline
 Bottleneck & FC 1 & 384 & - & - & - & -\\
 & FC 2 & 192 & - & - & - & -\\
 & FC 3 & 10 &  - & - & - & -\\
  & FC 4 & 192 & - & - & - & -\\
 & FC 5 & 384 & - & -  & - & -\\
 \hline
 Decoder & Deconv 1 & 32 & 8x8 & 5 & 2 & ReLU\\
 & Deconv 2 & 16 & 4x4 & 2 & 1 & ReLU\\
 & Deconv 3 & 8 & 4x4 & 2 & 1 & ReLU\\
 & Deconv 4 & 1 & 4x4 & 2 & 1 & Sigmoid\\
 \hline
\end{tabular}
\end{table}

Channel separation increases the use of computational resources compared to standard convolution approach, because it essentially uses three separate models: even though they are identical, they do not share weights. The number of model parameters is about three times that of the standard approach. Therefore, it requires roughly three times more computational power than the standard approach. Nonetheless, channel separation excels at distinguishing the object colours.

\subsection{Sampling and Binding}
Stochastic Gradient Variational Bayes-based sampling (SGVB) \cite{kingma2014auto} enables one-to-many mapping between action and language. The two VAEs have identical random sampling procedures. After producing the latent variables $z_{\text{mean}}$ and $z_{\text{var}}$ via the fully connected layers, we utilise a normal distribution $\mathcal{N}(\mu, \sigma^{2})$ to derive random values, $\epsilon$, which are, in turn, used with $z_{\text{mean}}$ and $z_{\text{var}}$ to arrive at the latent representation $z$, which is also known as the reparameterisation trick \cite{kingma2014auto}:
\begin{equation*}
    z =  z_{\text{mean}} + z_{\text{var}} \cdot \epsilon
\end{equation*}
where $\epsilon$ is the approximation of $\mathcal{N}(0, 0.01)$. 

As in the case of \cite{yamada2018paired}, to align the latent representations of robot actions and their descriptions, we use an extra loss term that brings the mean hidden features, $z_{\text{mean}}$, of the two VAEs closer to each other. This enables bidirectional translation between action and language, i.e., the network can transform actions to descriptions as well as descriptions to actions, after training without an explicit fusion of the two modalities. This loss term (binding loss) can be calculated as follows:
\begin{align*}
&L_{\text{binding}} = \sum_{i}^{B}\psi(z_{{\text{mean}}_{i}}^{\text{lang}}, z_{{\text{mean}}_{i}}^{\text{act}})+\sum_{i}^{B}\sum_{j\neq i}\text{max}\nonumber\\ &\left \{  0, \Delta + \psi(z_{{\text{mean}}_{i}}^{\text{lang}}, z_{{\text{mean}}_{i}}^{\text{act}}) - \psi(z_{{\text{mean}}_{j}}^{\text{lang}}, z_{{\text{mean}}_{i}}^{\text{act}})\right \},
\end{align*}
where $B$ stands for the batch size and $\psi$ is the Euclidean distance. The first term in the equation binds the paired instructions and actions, whereas the second term separates unpaired actions and descriptions. Hyperparameter $\Delta$ is used to adjust the separation margin for the second term - the higher it is, the further apart the unpaired actions and descriptions are pushed in the latent space.

Different multi-modal fusion techniques like Gated Multimodal Unit (GMU) \cite{arevalo2020gated}, which uses gating and multiplicative mechanisms to fuse different modalities, and CentralNet \cite{centralnet2018}, which fuses information by having a separate network for each modality as well as central joint representations at each layer, were also considered during our work. However, since our model is bidirectional (must work on both action-to-language and language-to-action directions) and must work with either language or action input during inference (both GMU and CentralNet require all of the modalities to be available), we opted for the binding loss for multi-modal integration.

\subsection{Loss Function}
The overall loss is calculated as the sum of the reconstruction, regularisation and binding losses. The binding loss is calculated for both VAEs jointly. In contrast, the reconstruction and regularisation losses are calculated independently for each VAE. Following \cite{yamada2018paired}, the reconstruction losses for the language VAE (cross entropy between input and output words) and action VAE (Euclidean distance between original and generated joint values) are $L_{\text{lang}}$ and $L_{\text{act}}$, respectively: 
\begin{align*}
L_{\text{lang}} &= \frac{1}{N-1} \sum_{t=1}^{N-1}\left ( -\sum_{i=0}^{V-1}  x_{t+1}^{\left [ i \right ]}\log y_{t}^{\left [ i \right ]}\right), \\
L_{\text{act}} &= \frac{1}{M-1} \sum_{t=1}^{M-1}\left \| j_{t+1} - \hat{\jmath}_{t+1}  \right \|_{2}^{2},
\end{align*}
where $V$ is the vocabulary size, $N$ is the number of words per description, M is the sequence length for an action trajectory. The regularisation loss is specific to variational autoencoders; it is defined as the Kullback–Leibler divergence for language $D_{{\text{KL}}_{\text{lang}}}$ and action $D_{{\text{KL}}_{\text{act}}}$. Therefore, the overall loss function is as follows:
\begin{equation*}
L_{\text{all}} = \alpha L_{\text{lang}} + \beta L_{\text{act}} + \gamma L_{\text{binding}} + \alpha D_{{\text{KL}}_{\text{lang}}} + \beta D_{{\text{KL}}_{\text{act}}}
\end{equation*}
where $\alpha$, $\beta$ and $\gamma$ are weighting factors for different terms in the loss function. In our experiments, $\alpha$ and $\beta$ are set to 1, whilst $\gamma$ is set to 2 in order to sufficiently bind the two modalities. 

\subsection{Transformer-Based Language Encoder}
In order for the model to understand unconstrained language input from non-expert human users, we replace the LSTM for the language encoder with a pretrained BERT-base language model \cite{devlin2019bert} - see Figure \ref{fig:architecture}. According to \cite{devlin2019bert}, BERT is pretrained with the BooksCorpus, which involves 800 million words, and English Wikipedia, which involves 2.5 billion words. With the introduction of BERT as the language encoder, we assume that BERT can interpret action descriptions correctly in our scenario. However, since language models like BERT are pretrained exclusively on textual data from the internet, they are not specialised for object manipulation environments like ours. Therefore, the embedding of an instruction like `\texttt{push the blue object}' may not differ from the embedding of another such as `\texttt{push the red object}' significantly. For this reason, we finetune the pretrained BERT-base, i.e. all of BERT's parameters are updated, during the end-to-end training of PVAE-BERT so that it can separate similar instructions from each other, which is critical for our scenario.

\subsection{Training Details}
To train the PVAE and PVAE-BERT, we first extract visual features using our channel-separated CAE. The visual features are used to condition the actions depending on the cube arrangement, i.e., the execution of a description depends also on the position of the target cube. For both the PVAE and PVAE-BERT, the action encoder and action decoder are each a two-layer LSTM with a hidden size of 100, whilst the language decoder is a single-layer LSTM with the same hidden size. In contrast, the language encoder of PVAE-BERT is the pretrained BERT-base model with 12 layers, each with 12 self-attention heads and a hidden size of 768, whereas the language encoder of the PVAE is a one-layer LSTM with a hidden size of 100. Both the PVAE and PVAE-BERT are trained end-to-end with both the language and action VAEs together. The PVAE and PVAE-BERT are trained for 20,000 and 40,000 iterations, respectively, with the gradient descent algorithm and Adam optimiser \cite{kingma2015adam}. We take the learning rate as $10^{-4}$ with a batch size of 100 pairs of language and action sequences after a few trials with different learning rates and batch sizes. Due to having approximately 110M parameters, compared with the PVAE's approximately 465K parameters, an iteration of PVAE-BERT training takes about 1.4 times longer than an iteration of PVAE training. Therefore, it takes about 2.8 times longer to train PVAE-BERT in total.

\section{Evaluation and Results}

\begin{table}[b]
\centering
\caption{Vocabulary}\label{tab:vocab}
\renewcommand{\tabcolsep}{10pt}
\begin{tabular}[t]{|c|c|c|}
\hline
& \textbf{Original} & \textbf{Alternative}\\
\hline
\textbf{Verb}&\texttt{push} & \texttt{move-up}\\
&\texttt{pull} & \texttt{move-down}\\
&\texttt{slide} & \texttt{move-sideways}\\
\hline
\textbf{Colour}&\texttt{red} & \texttt{scarlet} \\
&\texttt{green} & \texttt{harlequin}\\
&\texttt{blue} & \texttt{azure}\\
&\texttt{yellow} & \texttt{blonde}\\
&\texttt{cyan} & \texttt{greenish-blue} \\
&\texttt{violet} & \texttt{purple} \\
\hline
\textbf{Speed}&\texttt{slowly} & \texttt{unhurriedly}\\
&\texttt{fast} & \texttt{quickly} \\
\hline
\end{tabular}
\end{table}

\begin{table*}[ht]
\centering
\caption{Action-to-Language Translation Accuracies at Sentence Level}\label{tab:results}
\renewcommand{\tabcolsep}{5pt}
\begin{tabular}[t]{|c|cc|cc|cc|}
\hline
 Method	 & \multicolumn{2}{c|}{Experiment 1a (3 colours)} &  \multicolumn{2}{c|}{Experiment 1b (6 colours)}  &  \multicolumn{2}{c|}{Experiment 1c (6 shapes)}\\
 & Training & Test  & Training & Test & Training & Test\\
 \hline
PRAE + regular CAE & 33.33 $\pm$ 1.31\% & 33.56 $\pm$ 3.03\% & 33.64 $\pm$ 1.13\% & 33.3 $\pm$ 0.98\% & 68.36 $\pm$ 2.12\% & 65.28 $\pm$ 2.45\% \\
PVAE + regular CAE & 66.6 $\pm$ 1.31\% & 65.28 $\pm$ 6.05\% & 69.60 $\pm$ 0.46\% & 61.57 $\pm$ 2.01\% & 80.71 $\pm$ 1.41\% & 73.15 $\pm$ 1.87\% \\
PVAE + channel-separated CAE & \textbf{100.00 $\pm$ 0.00\%} & \textbf{90.28 $\pm$ 4.61\%} & \textbf{100.00 $\pm$ 0.00\%} & \textbf{100.00 $\pm$ 0.00\%} & \textbf{95.99 $\pm$ 3.74\%} & \textbf{92.13 $\pm$ 2.83\%}\\
 
\hline
\end{tabular}
\end{table*}

We evaluate the performance of our PVAE and its variant using BERT, namely PVAE-BERT, with multiple experiments. First, we compare the original PVAE with PRAE \cite{yamada2018paired} in terms of action-to-language translation by conducting experiments of varying object colour options to display the superiority of variational autoencoders over regular autoencoders and the advantage of using the channel separation technique in visual feature extraction. Different object colour possibilities correspond to a different corpus and overall dataset size; the more object colour options there are, the larger both the vocabulary and the overall dataset become. Therefore, with these experiments, we also test the scalability of both approaches. In order to show the impact of channel separation on the action-to-language translation performance, we train our architecture with visual features provided by a regular CAE (no channel separation) as implemented in \cite{yamada2018paired}. These are \textbf{Experiment 1a} (with 3 cube colour alternatives: red, green, blue) and \textbf{Experiment 1b} (with 6 cube colour alternatives: red, green, blue, yellow, cyan, violet) - see Table \ref{tab:results}. 

Moreover, in \textbf{Experiment 2}, we train PVAE-BERT on the dataset with 6 colour alternatives (red, green, blue, yellow, cyan, violet) to compare it with the standard PVAE by conducting action-to-language, language-to-language and language-to-action evaluation experiments. This experiment uses the pretrained BERT as the language encoder which is then finetuned with the rest of the model during training.

In Experiments 1a, 1b and 2, two cubes of different colours are placed on a table at which the robot is seated to interact with them. The words (vocabulary) that constitute the descriptions are given in Table \ref{tab:vocab}. We introduce a more diverse vocabulary by adding an alternative word for each word in the original vocabulary. As descriptions are composed of 3 words with two alternatives per word, we arrive at 8 variations for each description of a given meaning. Table \ref{tab:vocab} does not include nouns, because we use a predefined grammar, which doesn't involve a noun, and the same size cubes for these experiments.

For each cube arrangement, the colours of the two cubes always differ to avoid ambiguities in the language description. Actions, which are transcribed in capitals, are composed of any of the three action types PUSH, PULL, SLIDE, two positions LEFT, RIGHT and two speed settings SLOWLY, FAST, resulting in 12 possible actions ($3\text{ action types} \times 2 \text{ positions} \times 2\text{ speeds}$), e.g., PUSH-LEFT-SLOWLY means pushing the left object slowly. Every sentence is composed of three words (excluding the $<$BOS/EOS$>$ tags which denote \textsl{beginning of sentence} or \textsl{end of sentence}) with the first word indicating the action, the second the cube colour and the last the speed at which the action is performed (e.g., `\texttt{push green slowly}'). Therefore, without the alternative words, there are 18 possible sentences ($3\text{ action verbs} \times 3 \text{ colours} \times 2\text{ adverbs}$) for Experiment 1a, whereas, for Experiment 1b and 2, the number of sentences is 36 as 6 cube colours are used in both experiments. As a result, our dataset consists of 6 cube arrangements (3 colour alternatives and the colours of the two cubes on the table never match) for Experiment 1a, 12 cube arrangements for Experiments 1b and 2 (3 secondary colours are used in addition to 3 primary colours and secondary and primary colours are mutually exclusive), $18\times8=144$ possible sentences for Experiment 1a, $36\times8=288$ possible sentences for Experiments 1b and 2 with alternative vocabulary (consult Table \ref{tab:vocab}) - the factor of 8 because of eight alternatives per sentence. We have 72 patterns (action-description-arrangement combinations) for Experiment 1a (12 actions with six cube arrangements each) and 144 patterns for Experiments 1b and 2. Following Yamada et al. \cite{yamada2018paired}, we choose the patterns rigorously to ensure that combinations of action, description and cube arrangements used in the test set are excluded from the training set, although the training set includes all possible combinations of action, description and cube arrangements that are not in the test set. For Experiment 1a, 54 patterns are used for training while the remaining 18 for testing (for Experiments 1b and 2: 108 for training, 36 for testing). Each pattern is collected six times in the simulation with random variations on the action execution resulting in different joint trajectories. We also use 4-fold cross-validation to provide more reliable results (consult Table \ref{tab:results}) for Experiment 1.

\textbf{Experiment 1c} tests for different shapes, other than cubes: we perform the same actions on toy objects, which are a car, duck, cup, glass, house and lego brick. For testing the shape processing capability of the model, all objects are of the same colour, namely yellow. Analogous to the other experiments, two objects of different shapes are placed on the table. We keep the actions as they are but replace the colours with object names in the descriptions. Before we extract the visual features from the new images, we train both the regular CAE and the channel-separated CAE with them. Similar to Experiments 1a and 1b, we experiment with three methods: PRAE with standard CAE, PVAE with standard CAE and PVAE with channel-separated CAE. 

We use NICO (Neuro-Inspired COmpanion) \cite{kerzel2017nico, kerzel2020teaching} in a virtual environment created with Blender\footnote{https://www.blender.org/} for our experiments - see Figure \ref{fig:nico}. NICO is a humanoid robot, has a height of approximately one metre and a weight of approximately 20 kg. The left arm of NICO is used to interact with the objects while utilising 5 joints. Actions are realised using the inverse kinematics solver provided by the simulation environment: for each action, first, the starting point and endpoint are adjusted manually, then, the Gaussian deviation is applied around the starting point and endpoint to generate the variations of the action, ensuring that there is a slight difference in the overall trajectory. NICO has a camera in each of its eyes, which is used to extract egocentric visual images.

\subsection{Experiment 1}
We use the same actions as in \cite{yamada2018paired}, such as PUSH-RIGHT-SLOWLY. We use three colour options for the cubes as in \cite{yamada2018paired} for Experiment 1a, but six colours for Experiment 1b. However, we extend the descriptions in \cite{yamada2018paired} by adding an alternative for each word in the original vocabulary. Hence, the vocabulary size of 9 is extended to 17 for Experiment 1a and the vocabulary size of 11 is extended to 23 for Experiment 1b - note that we do not add an alternative for $<$BOS/EOS$>$ tags. Since every sentence consists of three words, we extend the number of sentences by a factor of eight ($2^{3}=8$).

After training the PVAE and PRAE on the same training set, we test them for action-to-language translation. We consider only those produced descriptions in which all three words and the $<$EOS$>$ tag are correctly predicted as correct. The produced descriptions that have one or more incorrect words are considered as false translations. As each description has seven more alternatives, predicting any of the eight description alternatives is considered correct. 

For Experiment 1a, our model is able to translate approximately 90\% of the patterns in the test set, whilst PRAE could translate only one third of the patterns, as can be seen in Table \ref{tab:results}. We can, thus, say that our model outperforms PRAE in one-to-many mapping. We also test the impact of channel separation on the translation accuracy by training our model with visual features extracted with the regular CAE as described in Yamada et al.'s approach \cite{yamada2018paired}. It is clearly indicated in Table \ref{tab:results} that using variational autoencoders instead of standard ones increases the accuracy significantly. Using PVAE with channel-separated CAE improves the results further, indicating the superiority of channel separation in our tabletop scenario. Therefore, our approach with variational autoencoders and a channel-separated CAE is superior to both PRAE and PVAE with regular visual feature extraction.

In Experiment 1b, in order to test the limits of our PVAE and the impact of more data with a larger corpus, we add three more colour options for the cubes: yellow, cyan and violet. These secondary colours are combined amongst themselves for the arrangements in addition to the colour combinations used in the first experiment, i.e., a cube of a primary colour and a cube of a secondary colour do not co-occur. Therefore, this experiment has 12 arrangements. Moreover, the vocabulary size is extended to 23 from 17 in Experiment 1b (two alternative words for each colour - see Table \ref{tab:vocab}). As in Experiment 1a, each sentence has eight alternative ways to be described.

We train both PVAE and PRAE \cite{yamada2018paired} on the extended dataset from scratch and test both architectures. As shown in Table \ref{tab:results} (Experiment 1b), PVAE succeeds in performing $100\%$ by translating every pattern from action to description correctly, even for the test set. In contrast, PRAE performs poorly in this setting and manages to translate only one third of the descriptions correctly in the test set. Compared with the accuracy values reached in the first experiment with less data and a smaller corpus, extension of the dataset helps PVAE to perform better in translation, whereas PRAE is not able to take advantage of more data. Similar to Experiment 1a, we also test the influence of channel separation on the translation accuracy by training PVAE with visual features provided by a regular CAE. In this setting, PVAE only achieves around $61\%$ of accuracy in the test set. This highlights once again the importance of channel separation in visual feature extraction for our setup. Whilst the improvement by using our PVAE over PRAE is significant, further improvement is made by utilising the channel-separated CAE. 

In addition, as the results show in the last column of Table \ref{tab:results} (Experiment 1c), our PVAE with channel separation in visual feature extraction outperforms the other methods even when manipulated objects have different shapes. Although there is a slight drop in action-language translation performance, it is clear that the PVAE with the channel-separated CAE is able to handle different-shaped objects. The PRAE model performs slightly better than it does in the experiments with cubes of different colours. However, our variational autoencoders approach without channel separation improves the translation accuracy by approximately 8\%. The channel separation in visual feature extraction improves the results even more similar to Experiment 1a and Experiment 1b, which shows the robustness of the channel-separated CAE when processing different objects.

\subsection{Experiment 2}

In this experiment, we test the performance of PVAE-BERT on action-to-language, language-to-action and language-to-language translation. We use the same dataset as in Experiment 1b for a fair comparison with the original PVAE (LSTM language encoder). We thus use the same descriptions, which are constructed by using a verb, colour and speed from the vocabulary given in Table \ref{tab:vocab} as well as the $<$BOS/EOS$>$ tags in the same order. Both PVAE and PVAE-BERT utilise channel-separated CAE-extracted visual features.

\begin{figure*}[ht]
    \centering
    \includegraphics[width=1\textwidth]{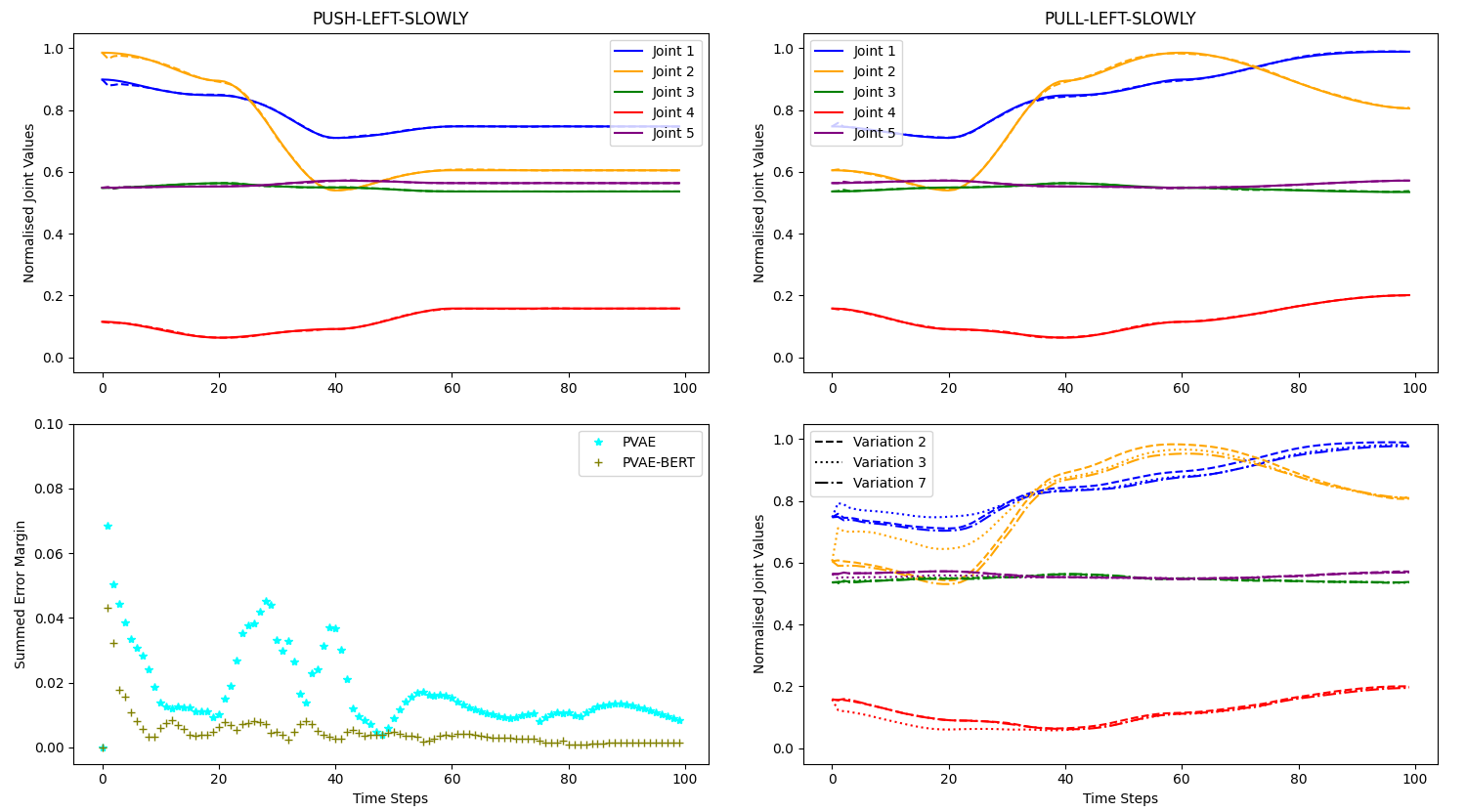}
    \caption[Joint angle trajectories]{Examples of language-to-action translation by PVAE-BERT and its comparison with PVAE: in the top row, the two plots represent the ground truth and predicted joint trajectories by PVAE-BERT for PUSH-LEFT-SLOWLY and PULL-LEFT-SLOWLY actions. Solid lines show the ground truth, while the dashed lines, which are often covered by the solid lines, show the predicted joint angle values. In the bottom row, the left plot shows the total error margin of the five joint values produced by PVAE and PVAE-BERT per time step for the PUSH-LEFT-SLOWLY action, while the right plot shows the joint values produced by PVAE-BERT given three variations (see Table \ref{tab:descvar}) of the same command for PULL-LEFT-SLOWLY - notice how the joint trajectories overlap most of the time. In all of the plots, the X axis represents the time steps.
    }
    \label{fig:joint}
\end{figure*}

\begin{table}[h]
\centering
\caption{Sentence Translation Accuracies for PVAE-BERT and PVAE}\label{tab:results2}
\begin{tabular}[t]{|c|c|c|}
\hline
  & PVAE & PVAE-BERT \\
 Translation Direction & Test Accuracy (T - F) & Test Accuracy (T - F)\\
 \hline
Action\textrightarrow Language & 100.00\% (216 - 0) & 97.22\% (210 - 6) \\
Language\textrightarrow Language & 100.00\% (216 - 0) & 80.56\% (174 - 42)\\
 \hline
\end{tabular}
\end{table}

As shown in Table \ref{tab:results2}, when translating from action to language, PVAE-BERT achieves approximately $97\%$ accuracy failing to translate only six of the descriptions, which is comparable with the original architecture - the original PVAE correctly translate all 216 descriptions. The false translations are all due to incorrect translation of cube colours, e.g., the predicted description is `\texttt{slide blue slowly}' instead of the ground truth `\texttt{slide red slowly}'. We hypothesise that the slight drop in the performance is due to the relatively small size of the dataset compared to almost 110 million parameters trained in the case of BERT. Nevertheless, these results show that finetuning the BERT during training leads to almost perfect action-to-language translation in our scenario.

As can be seen in Figure \ref{fig:joint}, both the PVAE and PVAE-BERT perform decently in language-to-action translation, and produce joint angle values that are in line with and very similar to the original descriptions. In the bottom left plot, we can see that the joint trajectories output by the PVAE-BERT are more accurate than those produced by the PVAE. We hypothesise that the error margins are negligible and both, PVAE-BERT and the PVAE, succeed in language-to-action translation. Since we did not realise the actions with the generated joint values in the simulation, we do not report the language-to-action translation accuracies in Table \ref{tab:results2}. However, we calculated the mean squared errors (MSE) for both the PVAE and PVAE-BERT, which were both very close to zero. Therefore, it is fair to say that both architectures recognise language and translate it to action successfully.

\begin{table}[h]
\centering
\caption{Variations of Descriptions for one Example and PVAE-BERT Language-to-Language Sentence Translation Accuracies}\label{tab:descvar}
\renewcommand{\tabcolsep}{1.5pt}
\scriptsize
\begin{tabular}[t]{|c|l|p{40mm}|c|}
\hline
\textbf{Var.} & \textbf{Type} & \textbf{Example} & \textbf{Accuracy}\\
\hline
1 & Standard & `\texttt{push blue slowly}' & 80.56\% \\
 \hline
2 & Changed Word Order & `\texttt{slowly push blue}' & 80.56\% \\
 \hline
3 & Full Command & `\texttt{push the blue cube slowly}' & 81.02\% \\
 \hline
4 & `please'+Full Command & `\texttt{please push the blue cube slowly}' & 81.94\% \\
 \hline
5 & Full Command+`please' & `\texttt{push the blue cube slowly please}' & 81.48\% \\
 \hline
6 & Ch.W.Order+F. Com.+`pls.' & `\texttt{slowly push the blue cube please}' & 81.48\% \\
 \hline
7 & Polite Request & `\texttt{could you please push the blue cube slowly?}' & 79.63\% \\
 \hline
\end{tabular}
\end{table}

Language-to-language translation, however, suffers a bigger performance drop when BERT is used as language encoder; PVAE-BERT reconstructs around $80\%$ of the descriptions correctly (see Table \ref{tab:results2}). We hypothesise that this is partly due to having an asymmetric language autoencoder with a BERT encoder and an LSTM decoder. The BERT-base language encoder constitutes the overwhelming majority of parameters in the PVAE-BERT model, which renders the language VAE heavily skewed to the encoder half. This may affect the performance of the language decoder when translating back to the description from the hidden code produced mainly by BERT as the decoder's parameters constitute less than 1\% of the parameters of the language VAE. This hypothesis is further supported by the original architecture, which has a symmetric language VAE, achieving $100\%$ of accuracy in the same task.

Nevertheless, our findings shows that the PVAE-BERT model achieves stable language-to-language translation performance even when the given descriptions do not comply with the fixed grammar and are full commands such as `\texttt{push the blue cube slowly}' or have a different word order such as `\texttt{quickly push blue}'. To turn predefined descriptions into full commands, we add the words `\texttt{the}' and `\texttt{cube}' to the descriptions and we also experiment with adding the word `\texttt{please}' and changing the word order as can be seen from the examples given in Table \ref{tab:descvar}. Although it is not explicitly stated in the table for space reasons, we alternate between the main elements of the descriptions as in the other experiments following the vocabulary; for example, `\texttt{push}' can be replaced by `\texttt{move-up}' and `\texttt{quickly}' can be replaced by `\texttt{fast}'. Moreover, we achieve consistent language-to-action translation performance with PVAE-BERT when we test it with different description types shown in the table - consult Figure \ref{fig:joint} bottom right plot. As the PVAE-BERT performs consistently even with descriptions not following the predefined grammar, we can see that the adoption of a language model to the architecture is promising towards acquiring natural language understanding skills.

\begin{figure*}[t]
    \centering
    \includegraphics[width=1\textwidth]{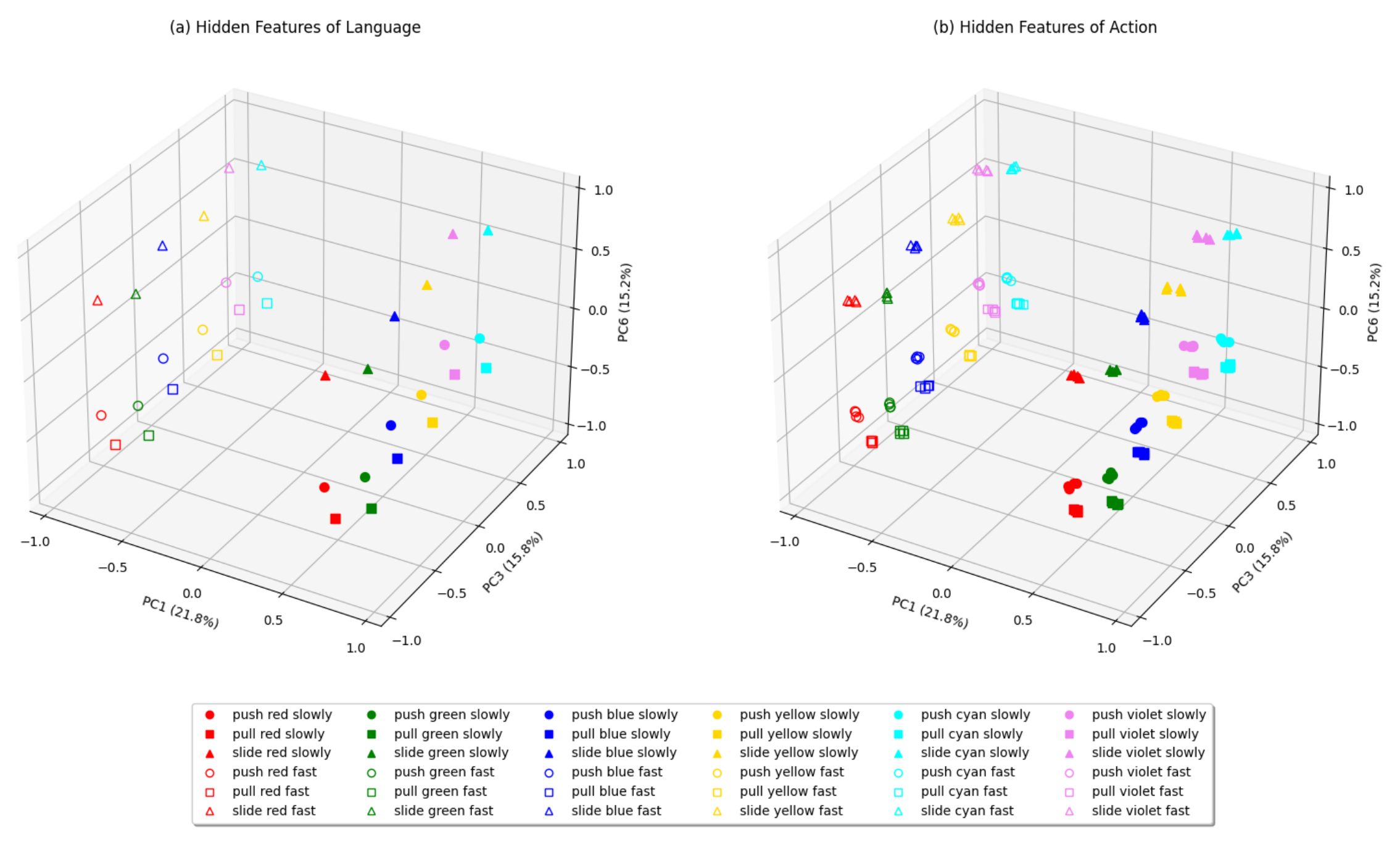}
    \caption[Hidden Features]{Hidden features of language (a) and hidden features of action (b): PCA was performed jointly on the hidden features of 36 descriptions and the hidden features of 144 actions. For (b), each unique action (12 in total) occurs 12 times as there are 12 possible cube arrangements; therefore, 144 points are shown. For both (a) and (b), we label the points according to descriptions, i.e., for (b), actions are also labelled according to their paired descriptions. As can be seen from the legend, different shapes, colours and fillings indicate the verb (action type), object colour and adverb (speed), respectively.
    }
    \label{fig:pca}
\end{figure*}

\subsection{Principal Component Analysis on Hidden Representations}

We have also conducted principal component analysis (PCA) on the hidden features extracted from PVAE-BERT. Figure \ref{fig:pca} shows the latent representations of language in Plot (a) and of action in Plot (b). The PCA on the representations of language shows that the model learns the compositionality of language: the X-axis (principal component PC~1) distinguishes the descriptions in the speed component (adverb), the Y-axis (PC~3) distinguishes colour, and the Z-axis (PC~6) distinguishes the action type (verb)\footnote{The percentages of variance explained were very similar between PC~2 until PC~6; therefore, we selected PC~3 and PC~6 for display as they resolved colour and action type optimally.}. Plot (b) shows that the PCA representations of actions are semantically similar, since their arrangement coincides with those in Plot (a).

Our method learns actions according to their paired descriptions: it learns the colour of the object (an element of descriptions) interacted with. However, it does not learn the position of it (an element of actions). We inspected the representations along all major principle components, but we could not find any direction along which the position was meaningfully distinguished. For example, in (b), some of the filled red circles (corresponding to description `\texttt{push red slowly}') are paired with the action PUSH-LEFT-SLOWLY while the others with PUSH-RIGHT-SLOWLY. As actions learned according to their paired descriptions, hence semantically, the filled red circles are grouped together even though the red cube may be on the right or left. In contrast, an action can be represented far from another identical action: e.g., 
the representations of `\texttt{pull red slowly}' (filled red circles in Figure~\ref{fig:pca}) are separated from those of `\texttt{pull yellow slowly}' (filled yellow circles) along PC~3, even if they both denote the action PULL-LEFT-SLOWLY. These results indicate that the binding loss has transferred semantically driven ordering from the language to the action representations.

When our agent receives a language instruction, which contains the colour but not position, the agent is still able to perform the action according to the position (cf.\ Figure~\ref{fig:joint}) of the object. The retrieval of the position information must therefore be done by the action decoder: it reads the images to obtain the position of the object that has the colour given in the instruction. It is therefore not surprising that the PCA does not reveal any object position encodings in the bottleneck.

\section{Discussion}
Experiments 1a and 1b show that our variational autoencoder approach with a channel-separated CAE visual feature extraction (‘PVAE + channel-separated CAE’) performs better than the standard autoencoder approach, i.e., PRAE \cite{yamada2018paired}, in the one-to-many translation of robot actions into language descriptions. Our approach is superior both in the case of three colour alternatives per cube and in the case of six colour alternatives per cube by a large margin. The additional experiment with six different objects highlights the robustness of our approach against the variation in object types. We demonstrate that a Bayesian inference-based method like variational autoencoders can scale up with more data for generalisation, whereas standard autoencoders cannot capitalise on a larger dataset, since the proposed PVAE model achieves better accuracy when the dataset and the corpus are extended with three extra colours or six different objects. Additionally, standard autoencoders are fairly limited in coping with the diversification of language as they do not have the capacity to learn the mapping between an action and many descriptions. In contrast, variational autoencoders yield remarkably better results in one-to-many translation between actions and descriptions, because stochastic generation (random normal distribution) within the latent feature extraction allows latent representations to slightly vary, which leads to VAEs learning multiple descriptions rather than a particular description for each action. 

A closer look into action-to-language translation accuracies achieved by the PRAE for Experiments 1a and 1b shows that having more variety in the data (i.e. more colour options for cubes) does not help the standard autoencoder approach to learn one-to-many binding between action and language. Both in the first case with three colour alternatives and in the second case with six colour alternatives, the PRAE manages to translate only around one third of the samples from actions to descriptions correctly. In contrast, the accuracies achieved by our proposed PVAE for both datasets prove that the variational autoencoder approach can benefit from more data as the test accuracy for the ‘PVAE + channel-separated CAE’ goes up by approximately $10\%$ to $100\%$ when three more colour options are added to the dataset.  

Furthermore, training the PVAE with the visual features extracted by the standard CAE demonstrates that training and extracting features from each RGB channel separately mitigates the colour distinction issue for cubes when the visual input, like in our setup, includes objects covering a relatively small portion of the visual field. The ‘PVAE + regular CAE’ variant performs significantly worse than our `PVAE + channel-separated CAE' approach. This also demonstrates the importance of the visual modality for the overall performance of the approach. Our analysis on the incorrectly translated descriptions shows that a large amount of all errors committed by the ‘PVAE + regular CAE’ were caused due to cube colour distinction failures such as translating `\texttt{slide red fast}' as `\texttt{slide green fast}', which proves the channel-separated CAE's superiority over the standard CAE in visual feature extraction in our scenario. Moreover, using the channel-separated CAE for visual feature extraction rather than the standard CAE results in better action-to-language translation accuracy even when the objects are of various shapes. This indicates that the channel-separated CAE not only works well with cubes of different colours but also objects of different shapes. We emphasise the superiority of channel separation in our scenario, which is tested and proven in a simulation environment. For real-world scenarios with different lighting conditions, it is advisable to take into account also the channel interaction \cite{tran2019channelseparated} to have more robust visual feature extraction.

Experiment 2 indicates the potential of utilising a pretrained language model like BERT for the interpretation of language descriptions. This extension produces comparable results to the original PVAE with the LSTM language encoder in language-to-action and action-to-language translations. The drop in language-to-language performance to $80\%$ is most probably caused by the asymmetric language VAE of the PVAE-BERT model that consists of a feedforward BERT encoder with attention mechanisms, which reads the entire input sequence in parallel, and of a recurrent LSTM decoder, which produces the output sequentially. A previous study on a text classification task also shows that LSTM models outperform BERT on a relatively small corpus because, with its large number of parameters, BERT tends to overfit when the dataset size is small \cite{ezen2020comparison}. Furthermore, we have also tested the PVAE-BERT, which was trained on predefined descriptions, with full sentence descriptions - e.g. `\texttt{push the blue cube slowly}' for `\texttt{push blue slowly}' - and with variations of the descriptions that have a different word order. We have confirmed that PVAE-BERT achieves the same performance in language-to-action and language-to-language translations. This is promising for the future because the pretrained BERT allows the model to understand unconstrained natural language commands that do not conform to the defined grammar.

The PCA conducted on the hidden features of PVAE-BERT shows that our method can learn language and robot actions compositionally and semantically. Although it is not explicitly given, we have also confirmed that both the PVAE and PVAE-BERT are able to reconstruct joint values almost perfectly accurately when we analysed the action-to-action translation results. Together with the language-to-language performance, action-to-action capability of both variants of our architecture demonstrates that the two variational autoencoders (language and action) in our approach retain their reconstructive nature.

\section{Conclusion}
In this study, we have reported the findings of previous work and its extension with several experiments. We have shown that variational autoencoders outperform standard autoencoders in terms of one-to-many translation of robot actions to descriptions. Furthermore, the superiority of our channel-separated visual feature extraction has been proven with an extra experiment that involves different types of objects. In addition, using the PVAE with a BERT model pretrained on large text corpora,
instead of the LSTM encoder trained on our small predefined grammar, unveils promising scaling-up opportunities for the proposed approach, and it offers the possibility to map unconstrained natural language descriptions with actions.

In the future, we will collect descriptions via crowdsourcing in order to investigate the viability of using a pretrained language model as an encoder to relate from language to motor control. We will also seek ways to bind the two modalities in a more biologically plausible way. Moreover, increasing the complexity of the scenario with more objects in general and on the table simultaneously may shed light to the scalability of our approach. Lastly, we will transfer our simulation scenario to the real world and conduct experiments on the real robot.

\section*{Acknowledgment}
The authors gratefully acknowledge support from the German Research Foundation DFG, project CML (TRR 169).

\bibliographystyle{plain}
\bibliography{references}

\begin{thebibliography}{10}

\bibitem{surveyxai}
Amina Adadi and Mohammed Berrada.
\newblock Peeking inside the black-box: A survey on explainable artificial
  intelligence (xai).
\newblock {\em IEEE Access}, 6:52138--52160, 2018.

\bibitem{Akakzia_DECSTR_2021}
Ahmed Akakzia, C{\'e}dric Colas, Pierre-Yves Oudeyer, Mohamed Chetouani, and
  Olivier Sigaud.
\newblock Grounding {{Language}} to {{Autonomously}}-{{Acquired Skills}} via
  {{Goal Generation}}.
\newblock In {\em International {{Conference}} on {{Learning
  Representations}}}, {Virtual (formerly Vienna, Austria)}, 2021.

\bibitem{antunes2019multitimescale}
Alexandre Antunes, Alban Laflaquiere, Tetsuya Ogata, and Angelo Cangelosi.
\newblock A bi-directional multiple timescales {LSTM} model for grounding of
  actions and verbs.
\newblock In {\em 2019 IEEE/RSJ International Conference on Intelligent Robots
  and Systems (IROS)}, pages 2614--2621, 2019.

\bibitem{arevalo2020gated}
John Arevalo, Thamar Solorio, Manuel Montes-y Gomez, and Fabio~A Gonz{\'a}lez.
\newblock Gated multimodal networks.
\newblock {\em Neural Computing and Applications}, 32(14):10209--10228, 2020.

\bibitem{bisk2020experience}
Yonatan Bisk, Ari Holtzman, Jesse Thomason, Jacob Andreas, Yoshua Bengio, Joyce
  Chai, Mirella Lapata, Angeliki Lazaridou, Jonathan May, Aleksandr Nisnevich,
  Nicolas Pinto, and Joseph Turian.
\newblock Experience grounds language.
\newblock In {\em Proceedings of the 2020 Conference on Empirical Methods in
  Natural Language Processing}, pages 8718--8735. Association for Computational
  Linguistics, November 2020.

\bibitem{chai2018language}
Joyce~Y Chai, Qiaozi Gao, Lanbo She, Shaohua Yang, Sari Saba-Sadiya, and
  Guangyue Xu.
\newblock Language to action: Towards interactive task learning with physical
  agents.
\newblock In {\em IJCAI}, pages 2--9, 2018.

\bibitem{chollet2017xception}
François Chollet.
\newblock Xception: Deep learning with depthwise separable convolutions.
\newblock In {\em 2017 IEEE Conference on Computer Vision and Pattern
  Recognition (CVPR)}, pages 1800--1807, 2017.

\bibitem{devlin2019bert}
Jacob Devlin, Ming-Wei Chang, Kenton Lee, and Kristina Toutanova.
\newblock {BERT}: Pre-training of deep bidirectional transformers for language
  understanding.
\newblock In {\em NAACL-HLT (1)}, 2019.

\bibitem{ELWW21}
Aaron Eisermann, Jae~Hee Lee, Cornelius Weber, and Stefan Wermter.
\newblock Generalization in multimodal language learning from simulation.
\newblock In {\em Proceedings of the International Joint Conference on Neural
  Networks (IJCNN 2021)}, Jul 2021.

\bibitem{ezen2020comparison}
Aysu Ezen-Can.
\newblock A comparison of {LSTM} and {BERT} for small corpus.
\newblock {\em arXiv preprint arXiv:2009.05451}, 2020.

\bibitem{hatori2018interactively}
Jun Hatori, Yuta Kikuchi, Sosuke Kobayashi, Kuniyuki Takahashi, Yuta Tsuboi,
  Yuya Unno, Wilson Ko, and Jethro Tan.
\newblock Interactively picking real-world objects with unconstrained spoken
  language instructions.
\newblock In {\em 2018 IEEE International Conference on Robotics and Automation
  (ICRA)}, pages 3774--3781. IEEE, 2018.

\bibitem{heinrich2018interactive}
Stefan Heinrich and Stefan Wermter.
\newblock Interactive natural language acquisition in a multi-modal recurrent
  neural architecture.
\newblock {\em Connection Science}, 30(1):99--133, 2018.

\bibitem{heinrich2020}
Stefan Heinrich, Yuan Yao, Tobias Hinz, Zhiyuan Liu, Thomas Hummel, Matthias
  Kerzel, Cornelius Weber, and Stefan Wermter.
\newblock Crossmodal language grounding in an embodied neurocognitive model.
\newblock {\em Frontiers in Neurorobotics}, 14:52, 2020.

\bibitem{hochreiter1997long}
Sepp Hochreiter and J{\"u}rgen Schmidhuber.
\newblock Long short-term memory.
\newblock {\em Neural Computation}, 9(8):1735--1780, 1997.

\bibitem{kerzel2020teaching}
Matthias Kerzel, Theresa Pekarek-Rosin, Erik Strahl, Stefan Heinrich, and
  Stefan Wermter.
\newblock Teaching {NICO} how to grasp: an empirical study on crossmodal social
  interaction as a key factor for robots learning from humans.
\newblock {\em Frontiers in Neurorobotics}, 14:28, 2020.

\bibitem{kerzel2017nico}
Matthias Kerzel, Erik Strahl, Sven Magg, Nicol{\'a}s Navarro-Guerrero, Stefan
  Heinrich, and Stefan Wermter.
\newblock {NICO}—{N}euro-{I}nspired {CO}mpanion: A developmental humanoid
  robot platform for multimodal interaction.
\newblock In {\em 2017 26th IEEE International Symposium on Robot and Human
  Interactive Communication (RO-MAN)}, pages 113--120, 2017.

\bibitem{kingma2015adam}
Diederik~P. Kingma and Jimmy Ba.
\newblock Adam: {A} method for stochastic optimization.
\newblock In {\em 3rd International Conference on Learning Representations,
  {ICLR}, San Diego, CA, USA, May 7-9}, 2015.

\bibitem{kingma2014auto}
Diederik~P. Kingma and Max Welling.
\newblock Auto-encoding variational {B}ayes.
\newblock In {\em Proceedings of International Conference on Learning
  Representations (ICLR), Banff, AB, Canada, April 14-16, 2014}, 2014.

\bibitem{lynch2021language}
Corey Lynch and Pierre Sermanet.
\newblock Language conditioned imitation learning over unstructured data.
\newblock {\em Robotics: Science and Systems}, 2021.

\bibitem{marino2021predictive}
Joseph Marino.
\newblock Predictive coding, variational autoencoders, and biological
  connections.
\newblock {\em Neural Computation}, 34(1):1--44, 2021.

\bibitem{ng2017hey}
Hwei~Geok Ng, Paul Anton, Marc Br{\"u}gger, Nikhil Churamani, Erik
  Flie{\ss}wasser, Thomas Hummel, Julius Mayer, Waleed Mustafa, Thi Linh~Chi
  Nguyen, Quan Nguyen, et~al.
\newblock Hey robot, why don't you talk to me?
\newblock In {\em 2017 26th IEEE International Symposium on Robot and Human
  Interactive Communication (RO-MAN)}, pages 728--731, 2017.

\bibitem{ogata2007parametricbias}
Tetsuya Ogata, Masamitsu Murase, Jun Tani, Kazunori Komatani, and Hiroshi~G.
  Okuno.
\newblock Two-way translation of compound sentences and arm motions by
  recurrent neural networks.
\newblock In {\em 2007 IEEE/RSJ International Conference on Intelligent Robots
  and Systems}, pages 1858--1863, 2007.

\bibitem{sak2014long}
Haşim Sak, Andrew Senior, and Françoise Beaufays.
\newblock {Long short-term memory recurrent neural network architectures for
  large scale acoustic modeling}.
\newblock In {\em Proceedings of Interspeech 2014}, pages 338--342, 2014.

\bibitem{shao2020concept2robot}
Lin Shao, Toki Migimatsu, Qiang Zhang, Karen Yang, and Jeannette Bohg.
\newblock Concept2robot: Learning manipulation concepts from instructions and
  human demonstrations.
\newblock In {\em Proceedings of Robotics: Science and Systems (RSS)}, 2020.

\bibitem{shridhar2018interactive}
Mohit Shridhar, Dixant Mittal, and David Hsu.
\newblock {INGRESS}: Interactive visual grounding of referring expressions.
\newblock {\em The International Journal of Robotics Research},
  39(2-3):217--232, 2020.

\bibitem{tran2019channelseparated}
Du~Tran, Heng Wang, Matt Feiszli, and Lorenzo Torresani.
\newblock Video classification with channel-separated convolutional networks.
\newblock In {\em 2019 IEEE/CVF International Conference on Computer Vision
  (ICCV)}, pages 5551--5560, 2019.

\bibitem{vaswani2017attention}
Ashish Vaswani, Noam Shazeer, Niki Parmar, Jakob Uszkoreit, Llion Jones,
  Aidan~N Gomez, {\L}ukasz Kaiser, and Illia Polosukhin.
\newblock Attention is all you need.
\newblock In {\em Advances in Neural Information Processing Systems}, pages
  5998--6008, 2017.

\bibitem{centralnet2018}
Valentin Vielzeuf, Alexis Lechervy, St{\'e}phane Pateux, and Fr{\'e}d{\'e}ric
  Jurie.
\newblock Centralnet: A multilayer approach for multimodal fusion.
\newblock In Laura Leal-Taix{\'e} and Stefan Roth, editors, {\em Computer
  Vision -- ECCV 2018 Workshops}, pages 575--589, Cham, 2019. Springer
  International Publishing.

\bibitem{wu2016google}
Yonghui Wu, Mike Schuster, Zhifeng Chen, Quoc~V Le, Mohammad Norouzi, Wolfgang
  Macherey, Maxim Krikun, Yuan Cao, Qin Gao, Klaus Macherey, et~al.
\newblock Google's neural machine translation system: Bridging the gap between
  human and machine translation.
\newblock {\em arXiv preprint arXiv:1609.08144}, 2016.

\bibitem{yamada2018paired}
Tatsuro Yamada, Hiroyuki Matsunaga, and Tetsuya Ogata.
\newblock Paired recurrent autoencoders for bidirectional translation between
  robot actions and linguistic descriptions.
\newblock {\em IEEE Robotics and Automation Letters}, 3(4):3441--3448, 2018.

\bibitem{yang-etal-2020-multilingual}
Yinfei Yang, Daniel Cer, Amin Ahmad, Mandy Guo, Jax Law, Noah Constant, Gustavo
  Hernandez~Abrego, Steve Yuan, Chris Tar, Yun-hsuan Sung, Brian Strope, and
  Ray Kurzweil.
\newblock Multilingual universal sentence encoder for semantic retrieval.
\newblock In {\em Proceedings of the 58th Annual Meeting of the Association for
  Computational Linguistics: System Demonstrations}, pages 87--94, Online, July
  2020. Association for Computational Linguistics.

\bibitem{Oezdemir_2021_ICDL}
Ozan Özdemir, Matthias Kerzel, and Stefan Wermter.
\newblock Embodied language learning with paired variational autoencoders.
\newblock In {\em 2021 IEEE International Conference on Development and
  Learning (ICDL)}, pages 1--6. IEEE, Aug 2021.

\end{thebibliography}

\begin{IEEEbiography}
[{\includegraphics[width=1in,height=1.25in,clip,keepaspectratio]{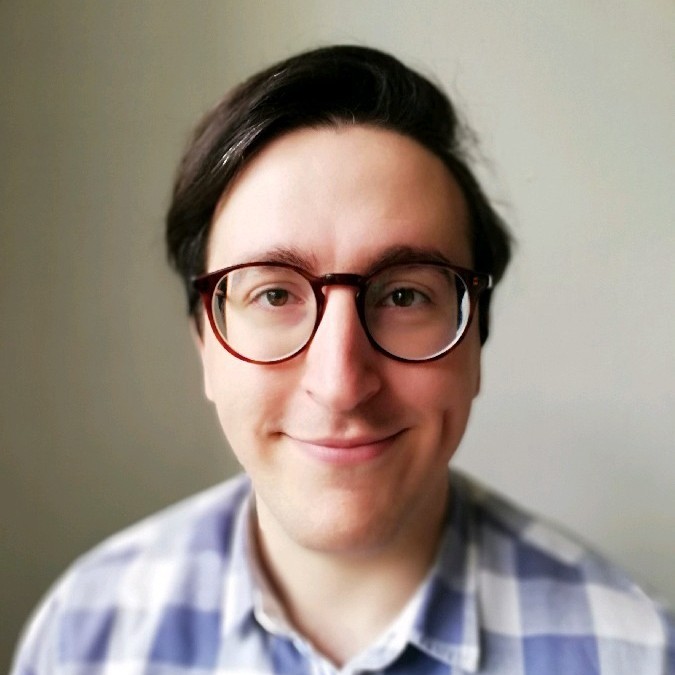}}]%
{Ozan \"Ozdemir} is a doctoral candidate and working as a research associate in the Knowledge Technology group, University of Hamburg, Germany. He has a BSc degree in computer engineering from Yildiz Technical University. He has received his MSc degree in Intelligent Adaptive Systems at the University of Hamburg. His research interests are embodied and crossmodal language learning, autoencoders, recurrent neural networks and large language models.
\end{IEEEbiography}
\vskip -2\baselineskip plus -1fil
\begin{IEEEbiography}
[{\includegraphics[width=1in,height=1.25in,clip,keepaspectratio]{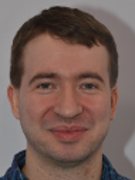}}]%
{Matthias Kerzel} received his MSc and PhD in computer science from the Universität Hamburg, Germany. He is currently a postdoctoral research and teaching associate at the Knowledge Technology Group of Prof. Stefan Wermter at the University of Hamburg. He has given lectures on Knowledge Processing in Intelligent Systems, Neural Networks and Bio-inspired Artificial Intelligence. He is currently the Secretary of the European Neural Network Society and worked in the organising committee of the International Conference on Artificial Neural Networks conferences. His research interests are in developmental neurorobotics, hybrid neurosymbolic architectures, explainable AI and human-robot interaction. He is currently involved in the international SFB/TRR-169 large-scale project on crossmodal learning.
\end{IEEEbiography}
\vskip -2\baselineskip plus -1fil
\begin{IEEEbiography}
[{\includegraphics[width=1in,height=1.25in,clip,keepaspectratio]{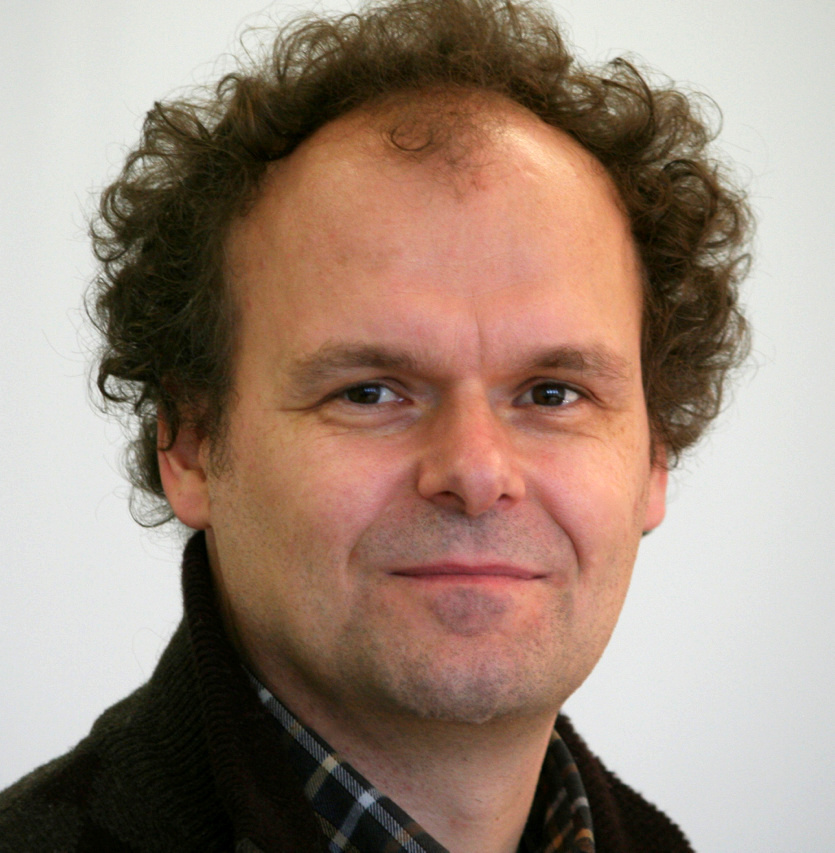}}]%
{Cornelius Weber} graduated in physics at Universität Bielefeld, Germany and received his PhD in computer science at Technische Universität Berlin. Following positions were a Postdoctoral Fellow in Brain and Cognitive Sciences, University of Rochester, USA; Research Scientist in Hybrid Intelligent Systems, University of Sunderland, UK; Junior Fellow at the Frankfurt Institute for Advanced Studies, Germany. Currently he is Lab Manager at Knowledge Technology, Universität Hamburg. His interests are in computational neuroscience, development of visual feature detectors, neural models of representations and transformations, reinforcement learning and robot control, grounded language learning, human-robot interaction and related applications in social assistive robotics.
\end{IEEEbiography}
\vskip -2\baselineskip plus -1fil
\begin{IEEEbiography}
[{\includegraphics[width=1in,height=1.25in,clip,keepaspectratio]{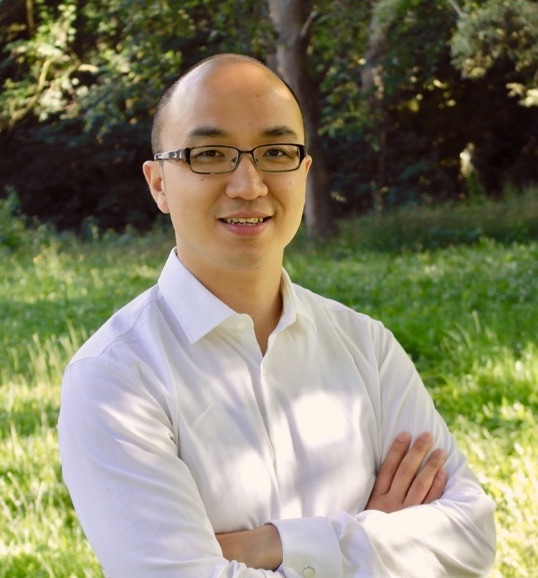}}]%
{Jae Hee Lee} is a postdoctoral research associate in the Knowledge Technology Group, University of Hamburg, Germany. He has worked on topics in multimodal learning, grounded language understanding, and spatial and temporal reasoning. Jae Hee Lee received his Diplom degree in mathematics and doctoral degree in computer science from the University of Bremen, Germany. He was a postdoctoral researcher at the Australian National University, University of Technology Sydney (Australia) and Cardiff University (UK).
\end{IEEEbiography}
\vskip -2\baselineskip plus -1fil
\begin{IEEEbiography}
[{\includegraphics[width=1in,height=1.25in,clip,keepaspectratio]{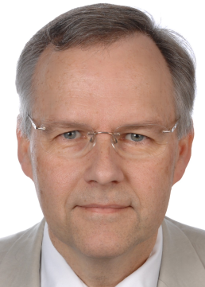}}]%
{Stefan Wermter} (Member, IEEE) is currently a Full Professor with the University of Hamburg, Hamburg, Germany, where he is also the Director of the Department of Informatics, Knowledge Technology Institute. Currently, he is a co-coordinator of the International Collaborative Research Centre on Crossmodal Learning (TRR-169) and a coordinator of the European Training Network TRAIL on transparent interpretable robots. His main research interests are in the fields of neural networks, hybrid knowledge technology, cognitive robotics and human–robot interaction. 
He is an Associate Editor of Connection Science and International Journal for Hybrid Intelligent Systems. He is on the Editorial Board of the journals Cognitive Systems Research, Cognitive Computation and Journal of Computational Intelligence. He is serving as the President for the European Neural Network Society.
\end{IEEEbiography}

\end{document}